# VIRUS-MNIST: MACHINE LEARNING BASELINE CALCULATIONS FOR IMAGE CLASSIFICATION

Erik Larsen, Korey MacVittie, John Lilly

PeopleTec, Inc., 4901 Corporate Drive. NW, Huntsville, AL, USA
erik.larsen@peopletec.com

## ABSTRACT

*The Virus-MNIST data set is a collection of thumbnail images that is similar in style to the ubiquitous MNIST hand-written digits. These, however, are cast by reshaping possible malware code into an image array. Naturally, it is poised to take on a role in benchmarking progress of virus classifier model training. Ten types are present: nine classified as malware and one benign. Cursory examination reveals unequal class populations and other key aspects that must be considered when selecting classification and pre-processing methods. Exploratory analyses show possible identifiable characteristics from aggregate metrics (e.g., the pixel median values), and ways to reduce the number of features by identifying strong correlations. A model comparison shows that Light Gradient Boosting Machine, Gradient Boosting Classifier, and Random Forest algorithms produced the highest accuracy scores, thus showing promise for deeper scrutiny.*

## 1. INTRODUCTION

Detecting cyber security threats has become more difficult in recent years due to the evolving nature of obfuscation techniques employed by malicious hackers. Vu et al [1] find that there are some 323,000 new malware file detections every day. Identifying malicious code will be essential in securing personal computers and networks as technology continues to evolve. Toward this endeavour, one detection technique explored by Nataraj et al. [2] transforms 25 different species of malware files into images. Herein we build upon the work of Noever and Noever [3] who extended the malware image problem into the realm of the familiar MNIST image classification task. They use nine virus classes and one benign class based on byte-similarities between the malicious files. Figure 1 shows random examples of each type of thumbnail created.

We report scoring results for 21 machine learning models trained on the Virus - MNIST dataset [4]. There are 51,880 total examples (48,422 training and 3,454 test), each containing 1,026 features. Two of these, "labels" and "hash", are provided with an assumed fidelity greater than 95% making supervised learning approaches viable. This leaves a total of 1,024 pixel values describing thumbnail-size, grayscale images of shape (32, 32, 1). These pictures, seen in Figure 1, resemble static noise seen on an old television. However, clear features are present to distinguish the examples. Darker and lighter regions as well as stripes of solid white or black are easily evident upon cursory inspection. The dimensions for height and width are ideal for immediate ingestion into a convolutional network with no pre-processing. No missing pixel values were detected. Train and test sets required 400.1 MB of storage.

## 2. METHODS

This project used both Kaggle and Google Colab notebooks with the Virus-MNIST dataset found at https://www.kaggle.com/datamunge/virusmnist along with individual Python IDEs [5-7]. ML libraries used include PyCaret, UMAP, SciKit-Learn, and TensorFlow's Keras API (2.4.0). Training and evaluation occur on a total of 21 classifiers. The given test set remains unseen by the algorithm and was only used for the final model scoring independent from PyCaret.

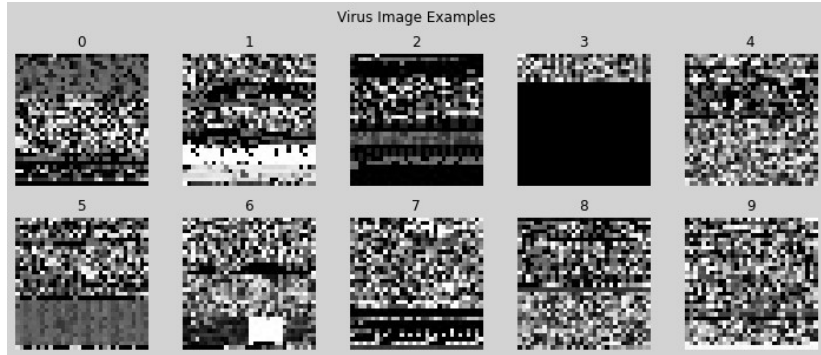

*Figure 1.* Samples from each class

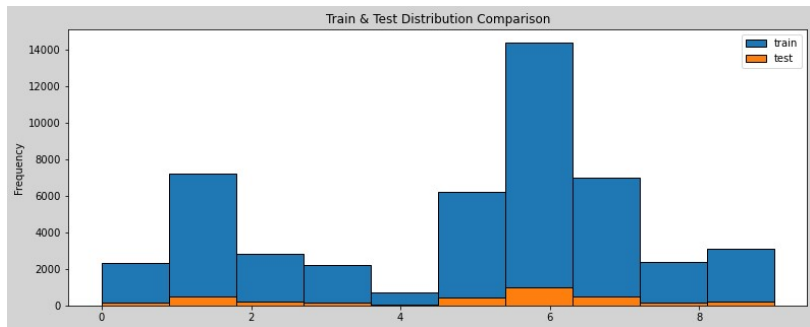

*Figure 2.* A comparison shows the relative sizes and similar distributions

All "hash" labels were string objects of uniform length. In a Pycaret general comparison, Light Gradient Boosting Machine (lightgbm), Gradient Boosting Classifier (gbc), and Random Forest (rf) are the top performing models with accuracy scores of 0.880, 0.8624, and 0.8574, respectively. These three are further optimized along with convolutional networks including known models such as AlexNet and LeNet-5 [8, 9]. The CatBoostClassifier and GradientBoostingClassifier also performed with similar accuracy and F1 scores but training takes too long to be viable with a data set of this size using either classifier.

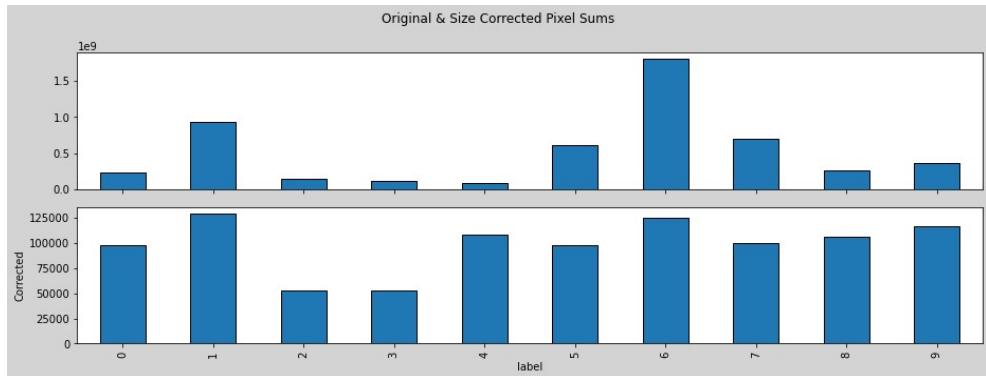

*Figure 3.* The corrected pixel sums show differences that help identify each class

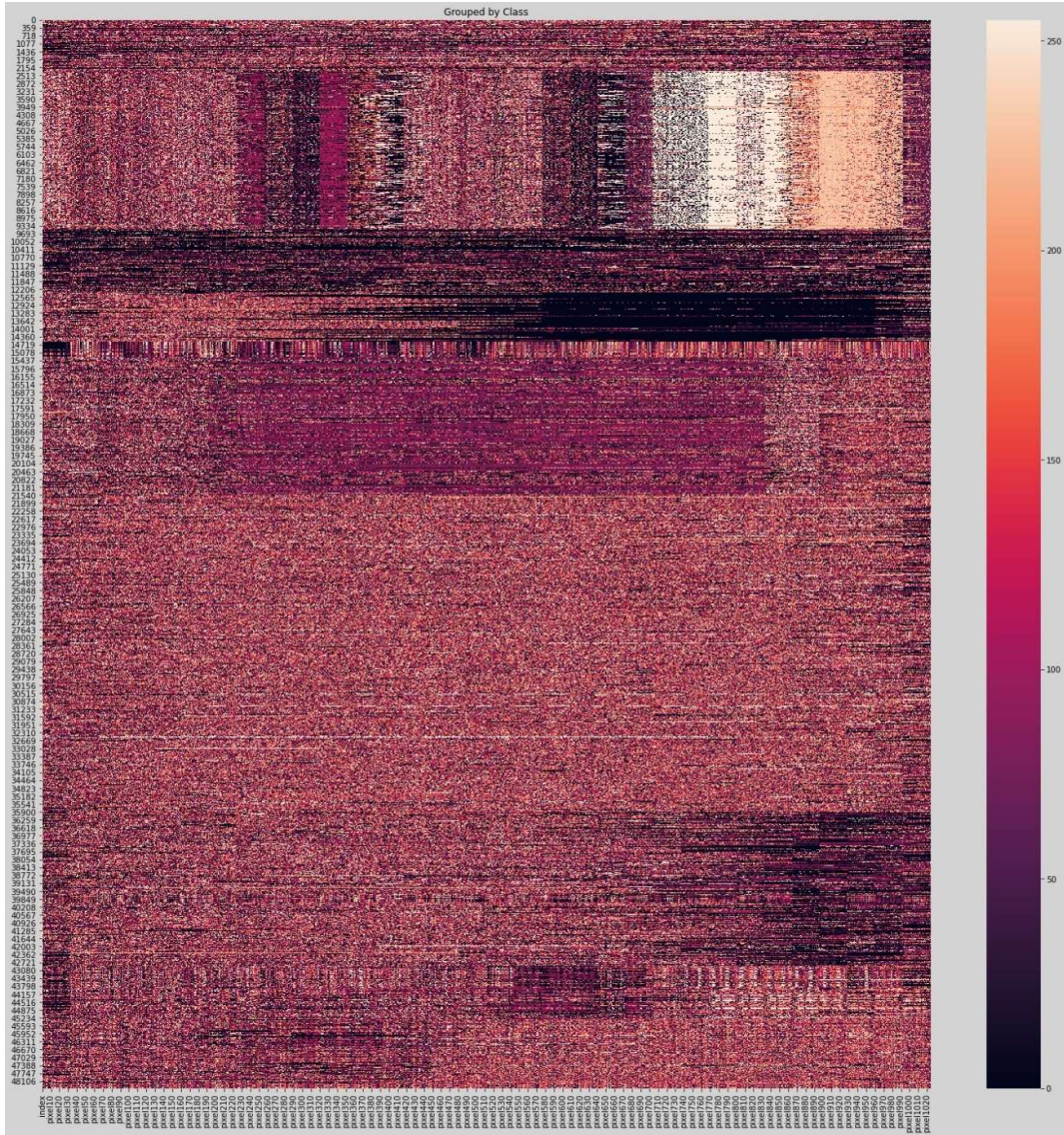

*Figure 4*. Heatmap of pixel values sorted by class

## 2.1. Data Preparation & EDA

Ten classes of "malware" are represented by these images, with one class known as benign for reference. The population across classes is not uniform, with class six having more than twice as many examples of any other as evidenced by Figure 2. The least represented is class four, with less than half as any other. The test-train ratio is 7.1% with corresponding population distributions, which is appropriate for the large amount of data. Pixel values range from 0 – 255, with each class showing position-based behaviour inside the image arrays. Figure 3 depicts the total sum of pixel values and the normalized values. This becomes exaggerated in aggregate statistics such as the median pixel value by class. The distinction is most easily seen using heatmaps, some of which are displayed in Figure 4 and Figure 5.

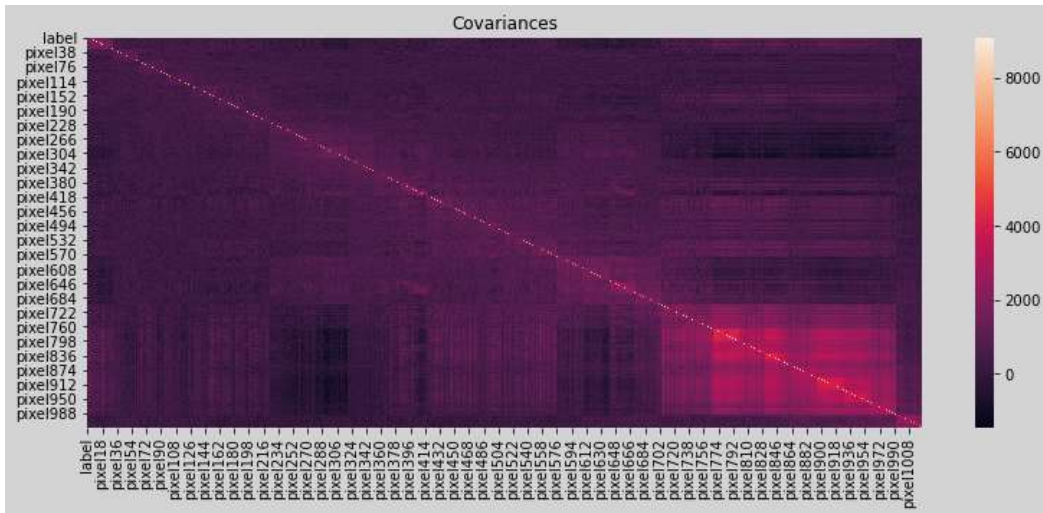

*Figure 5.* Heatmap of covariances between "features"

UMAP's unsupervised K-Means clustering showed poor performance in all pertinent metrics. This is demonstrated by the tight grouping in Figure 6 below. UMAP's inability to find separable hyper-plane clusters is shown in this 2-D projection.

EDA revealed regions of higher pixel intensity for many different statics, promising avenues for feature engineering. One example can be seen in Figure 7 which shows how the individual pixel sums in each class are different and could lead to visual discrimination. Figure 8 is similar, displaying a clear trend in standard deviation between pixel medians. Comparing such descriptive statistics across each class emphasizes their unique identities, suggesting future improvements for solution boundary hyper-plane separability and ease of classification, as well as faster processing time with the most suited algorithms. Observing pixel distributions side-by-side also demonstrates some identifiable consonance and dissonance between classes. Heatmaps for these statistical configurations, particularly the median, appear to give each class a spectral "fingerprint," or unique barcode-like identifier. This could possibly be exploited for non-critical situations, where time allows for more processing.

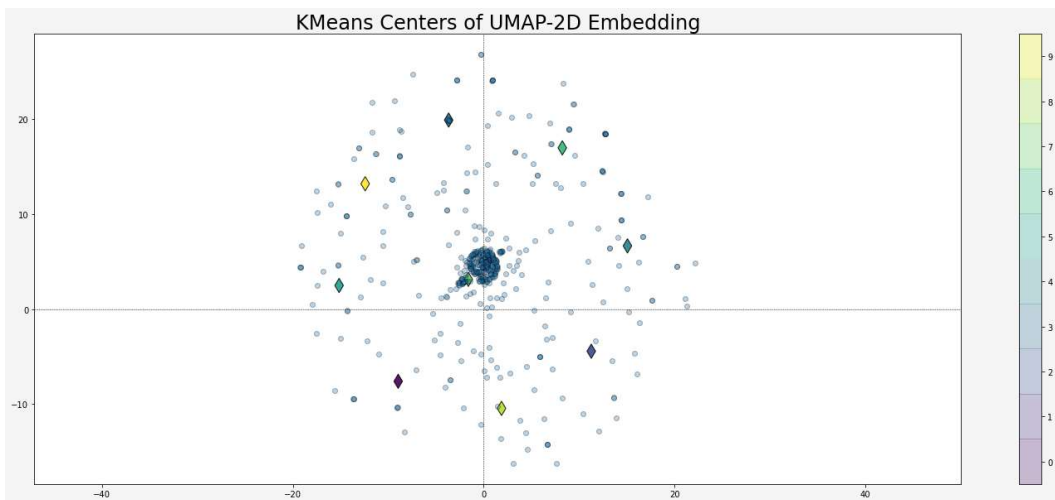

*Figure 6.* Poor separation leads to an F1 score of only 10.3%

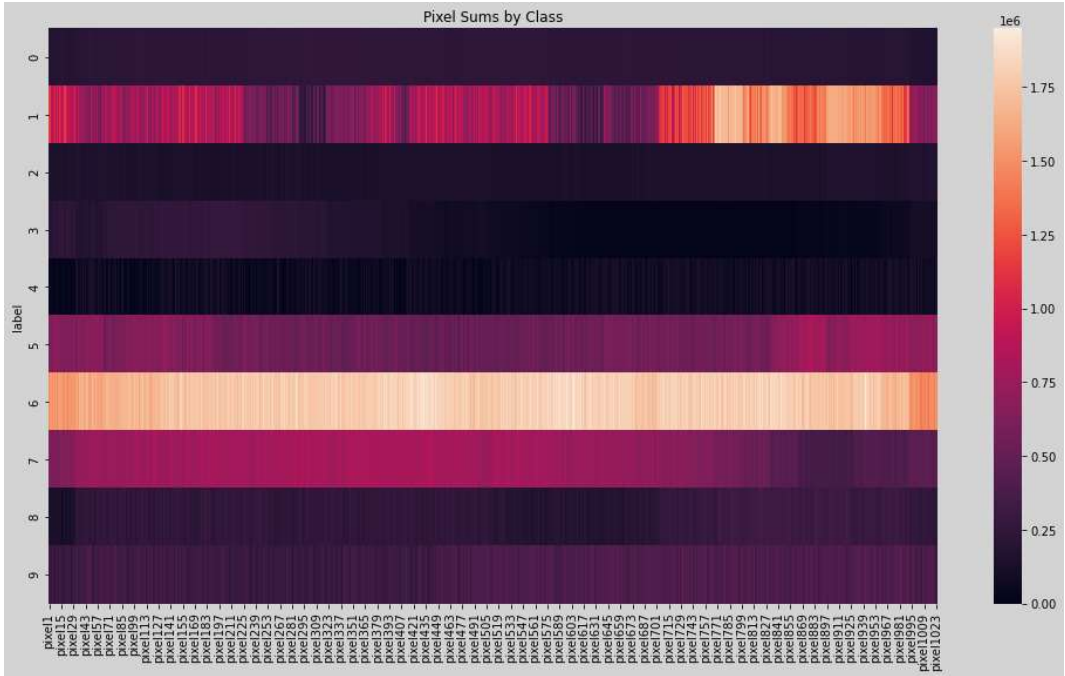

*Figure 7.* Heatmap showing different "signatures" or individual pixel sums in each class

## 2.2 Validation & Scoring

### 2.2.1 Pycaret

This high-level python library [10] compares the performance of multiple machine learning algorithms from a single platform. It must be downloaded and installed into the working environment using the pip package manager. For classification, 15 models are compiled and run with the chosen number of stratified cross-validation (cv) folds; in this case, cv = 2. The size and difficulty in achieving convergence slowed the total processing time to over 15 hours! Decreasing the number of folds (default cv = 10) is necessary to prevent time-out but has a negligible effect on the scores, and therefore, the rankings.

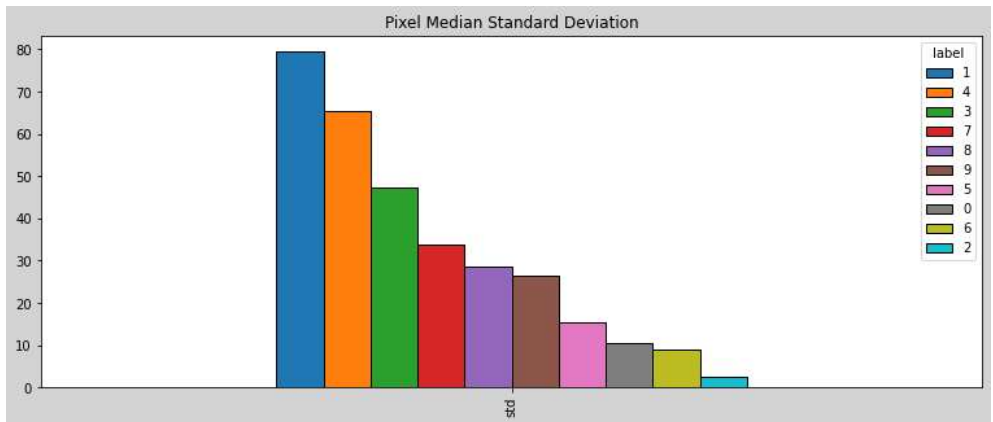

*Figure 8.* Standard deviation of pixel medians trend and grouping

Using this comparison, the top three performing models are Light Gradient Boosting Machine (lightgbm), Gradient Boosting Classifier (gbc), and Random Forest (rf). The details of eight models can be found in Table 1, while the rest are available in the notebooks indicated in the References section. Lightgbm's accuracy was 0.8800 but the other two trail by less than 3% at 0.8624 and 0.8574, respectively. However, gbc takes about five times as long to train as lightgbm, and over 100 times longer than rf (for a paltry 1% gain). Therefore, it seems that an accurate model could be trained using lightgbm and deployed—through transfer learning—to a rf classifier deployed on the edge where low classification time is critical. Table 1 displays the results for all 13 classifiers.

Table 1. Pycaret Model Comparison Results

| Model | Accuracy | AUC | F1 | TT (Sec) |
|---|---|---|---|---|
| Light Gradient Boosting Machine | 0.8800 | 0.9886 | .8758 | 221.715 |
| Gradient Boosting Classifier | 0.8624 | 0.9834 | 0.8605 | 10045.86 |
| Random Forest Classifier | 0.8574 | 0.9834 | 0.8514 | 73.445 |
| Extra Trees Classifier | 0.8518 | 0.9818 | 0.8428 | 45.320 |
| Logistic Regression | 0.8341 | 0.9710 | 0.8399 | 464.890 |
| Naïve Bayes | 0.8179 | 0.9704 | 0.8077 | 30.260 |
| SVM – Linear Kernel | 0.8130 | 0.0000 | 0.8260 | 40.310 |
| Linear Discriminant Analysis | 0.7987 | 0.9658 | .8018 | 41.470 |
| Ridge Classifier | 0.7090 | 0.0000 | 0.7087 | 28.275 |
| Decision Tree Classifier | 0.6850 | 0.8205 | 0.6943 | 81.650 |
| K Neighbors Classifier | 0.6634 | 0.8358 | 0.6470 | 892.000 |
| Quadratic Discriminant Analysis | 0.4176 | 0.6048 | 0.3162 | 54.565 |
| Ada Boost Classifier | 0.3154 | 0.7074 | 0.3099 | 229.755 |

The Kaggle notebook contains hyper-parameters used to achieve these results, but model optimization occurs with no prior knowledge of these to probe a more robust grid. An optimized lgbm classifier achieved accuracy and F1 scores of 91% and 92%, respectively, almost besting the CNN algorithms. These scores are shown in Table 2 along with the following models for comparison. The confusion matrix for PyCaret's Random Forest Classifier is shown in Figure 9. Figure 10 displays the change in performance and training time by applying different methods of regularization to remove high variance.

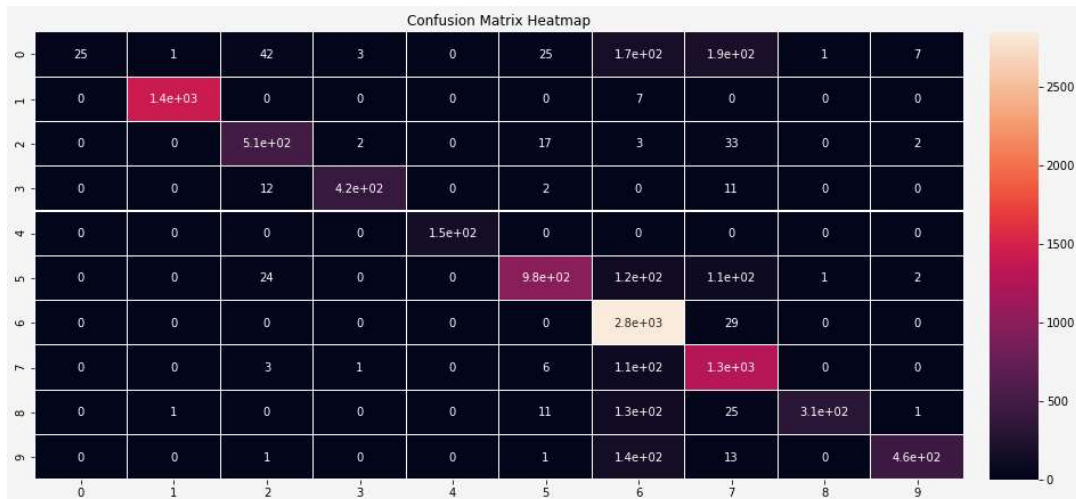

*Figure 9.* Random Forest results show greatest confusion in classes 6 and 7

### 2.2.2 CapsNet

We also developed a CapsNet (capsule network) [11] model trained with the data. The model was trained for 100 epochs on batch sizes of 64. It performed reasonably well on the training portion at 64% accuracy, but it only achieved 22% accuracy on the test data. Additional training or other modifications to the model, such as increasing the number of layers or nodes in layers or incorporating different activation functions (this model used RELU for hidden layers and sigmoid for the output layer), may enhance accuracy, as the training data accuracy indicates it is possible for this type of model to differentiate these classes. Conversely, the complex nature of capsule networks suggests it would be ineffective.

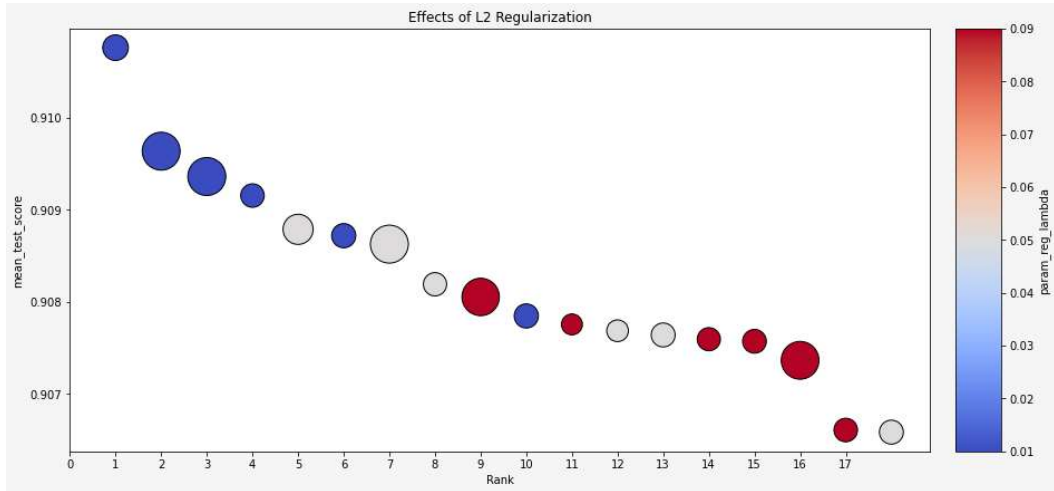

*Figure 10.* Results of applying 'lasso' regularization to lgbm, sized by fit time

### 2.2.3 Graph Convolutional Network

The adjacency graphs generated for each class were sufficiently differentiated, and the GCN model [12] achieved an accuracy of 90.6% on the test data after 80 epochs of training. The model had difficulty breaching into higher accuracy, stalling out in the low 90's for most of its training time. It is possible that GCN-based approaches to this dataset may not be able to exceed that, but also possible that a more sophisticated network—like one that incorporates elements of our CNN approach (see below)—may result in a more accurate model.

Table 2. CapsNet, CNN & GCN Results

| Model | Accuracy | Precision | Recall | F1 | TT (Sec) |
|---|---|---|---|---|---|
| CapsNet | .22 | .16 | .22 | .17 | – |
| Convolutional Neural Network | .94 | .94 | .90 | .89 | 626.3 |
| GCN | .906 | – | – | – | 1683 |

*Note:* The training time for this CapsNet model cannot be accurately reported due to interruptions of the machine by which the model was being trained.

### 2.2.4 Convolutional Neural Networks

The images' heights and widths are a power of two, requiring no resizing before ingestion. Initial runs showed the scoring performance for which CNNs are renowned despite being highly prone to overfitting on the virus images. Initial training scores reach as high as 96% while test scores are typically around 80%. Multiple strategies resolve high variance: adding dropout layers and regularization to the fully connect (FC) section more closely aligned train and test scores, and effectively restoring the model's ability to generalize to new data. The distribution of pixel means seen in Figure 11 can help these models distinguish between classes as well.

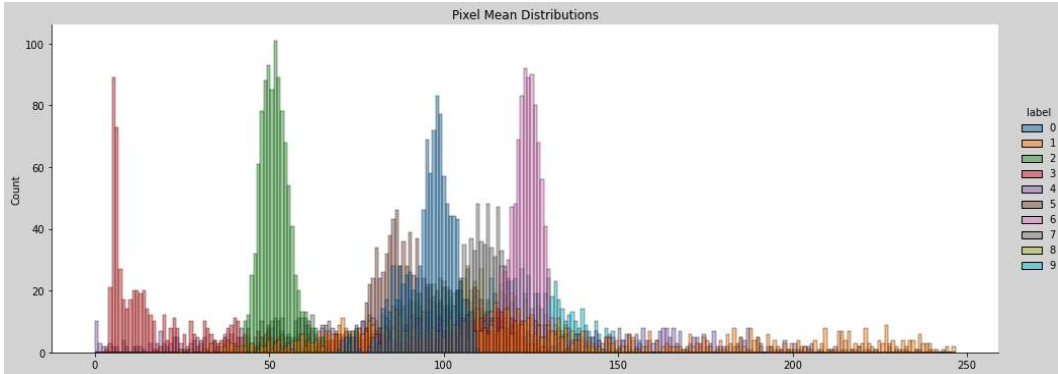

*Figure 11.* Individual pixel means show defining characteristics that help distinguish each class

## 3. RESULTS

We report results for a convolutional neural network created with minimal architecting that achieved training and test accuracies of approximately 93%. Care must be taken when examining accuracy because of the low number of class zero predictions. Overfitting has been essentially removed, and the potential exists to bring them near unity with a more sophisticated algorithmic approach. Notably, the classification report revealed relatively poor performance in precision and recall (hence F1 as well) for class zero compared to all others. Class zero is the known benign class suggesting that the malware classes have characteristics that are distinguishable by the network. Figure 12 shows that training above 50 epochs yields little return using a CNN with the LeNet-5 architecture, while its final results are seen in Figure 13.

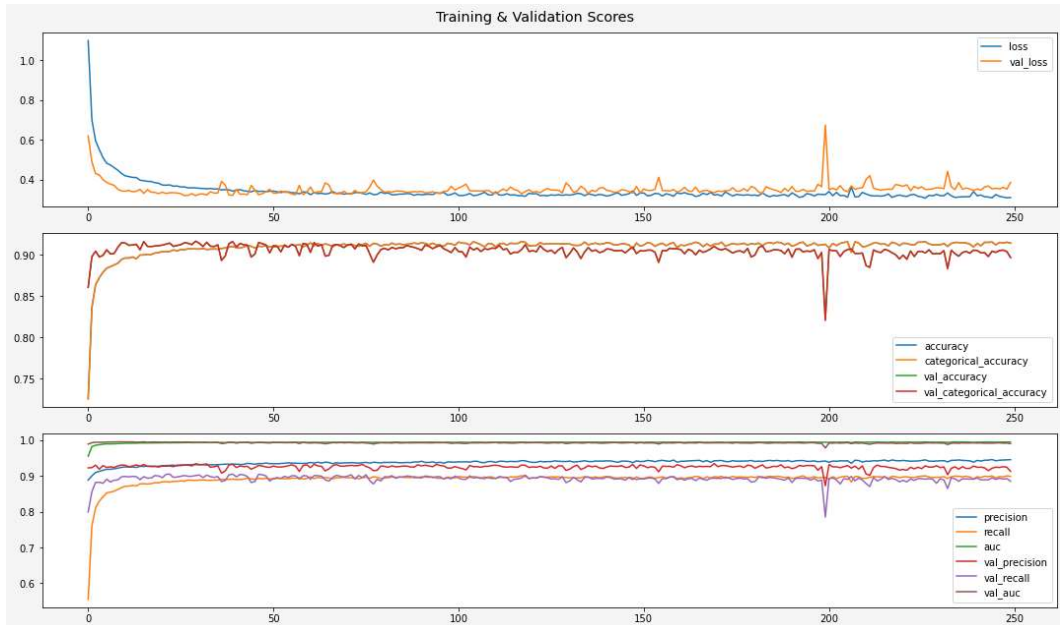

*Figure 12.* Results for a CNN with a modified LeNet-5 architecture

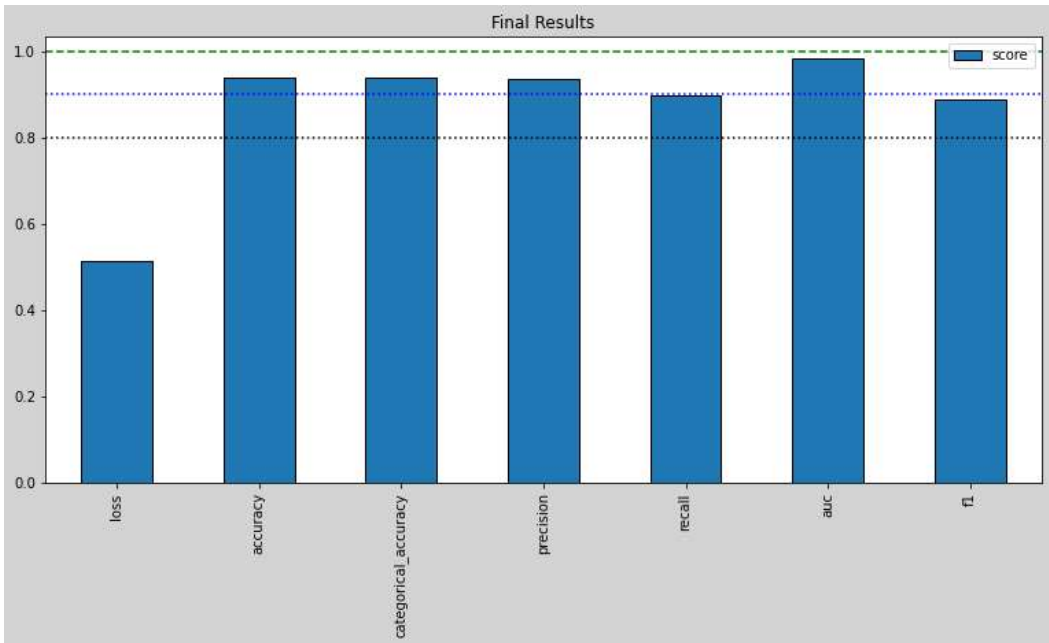

*Figure 13.* The final testing scores for modified LeNet-5

## 4. CONCLUSIONS

Many algorithms were able to break 90% accuracy and F1 scores. The convolutional method performed only slightly better than GCN and LGBM approaches. Image manipulations such as rotation and translation tend to improve these scores in object classification. However, physical objects and distances are invariant to these transformations. VMNIST thumbnails are not images of physical origin, therefore introducing such variations would likely confuse the algorithm and decrease performance, if not destroy the malware's identity completely.

## ACKNOWLEDGEMENTS

The authors would like to thank the PeopleTec, Inc. Technical Fellows for facilitating this research.

# Authors

Erik Larsen, M.S. is a senior data scientist specialized in deep learning and prior research in Density Functional Theory. He completed both M.S. and B.S. in Physics at the University of North Texas, and a B.S. in Professional Aeronautics from Embry-Riddle Aeronautical University while serving as an aviator in the U.S. Army.

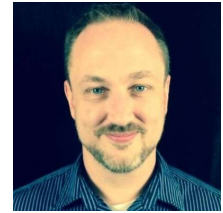

Korey MacVittie, M.S. is a data scientist specialized in machine learning. Prior Prior research includes identifying undervalued players in sports drafting. He completed his M.S. at Southern Methodist University, and a B.S. in Computer Science and B.S. in Philosophy from University of Wisconsin Green Bay.

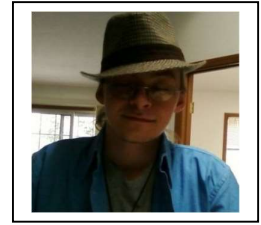

John R. Lilly III is an applied machine learning research scientist specializing in quantitative analysis, algorithm development, and implementation for embedded systems. Prior research is in token economics and the deployment of crypto assets across multiple enterprise blockchain protocols. He has been trained in statistics by Cornell University. Formerly infantry, he currently serves as an intelligence advisor for the U.S. Army's Security Force Assistance Brigade.

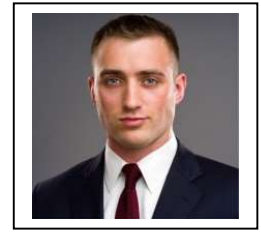